\pdfoutput=1

\documentclass[11pt]{article}

\usepackage[]{EMNLP2023}

\usepackage{times}
\usepackage{latexsym}

\usepackage[T1]{fontenc}

\usepackage[utf8]{inputenc}

\usepackage{microtype}

\usepackage{inconsolata}
\usepackage{float}
\usepackage{amsfonts}
\usepackage{pdfpages}

\usepackage{color}
\usepackage{algorithm}
\usepackage{algorithmic}
\usepackage{amsmath}
\usepackage{amsthm}
\usepackage{enumitem}
\usepackage{bm}
\usepackage{subfigure} 
\usepackage{nicefrac}
\usepackage{tabularx}
\usepackage{url}
\usepackage{booktabs}
\usepackage{multirow}
\usepackage{graphicx}
\usepackage{amssymb}
\usepackage{bbding}

\graphicspath{ {./images/} }

\newcolumntype{L}[1]{>{\raggedright\let\newline\\\arraybackslash\hspace{0pt}}m{#1}}
\newcolumntype{C}[1]{>{\centering\let\newline  \\\arraybackslash\hspace{0pt}}m{#1}}
\newcolumntype{R}[1]{>{\raggedleft\let\newline \\\arraybackslash\hspace{0pt}}m{#1}}

\title{Ensemble Learning for Graph Neural Networks}

\author{Zhen Hao Wong \quad Ling Yue \quad Quanming Yao \\
	Department of Electronic Engineering, Tsinghua University, Beijing, China \\
        \texttt{\small huang-zh20@mails.tsinghua.edu.cn, }
        \texttt{\small {lingyue/qyaoaa}@tsinghua.edu.cn}
}

\begin{document}
\maketitle
\begin{abstract}
Graph Neural Networks (GNNs) have shown success in various fields for learning from graph-structured data. 
    This paper investigates the application of ensemble learning techniques to improve the performance and robustness of Graph Neural Networks (GNNs). 
    By training multiple GNN models with diverse initializations or architectures, 
    we create an ensemble model named ELGNN that captures various aspects of the data and 
    uses the Tree-Structured Parzen Estimator algorithm to determine the ensemble weights. 
    Combining the predictions of these models enhances overall accuracy, 
    reduces bias and variance, and mitigates the impact of noisy data. 
    Our findings demonstrate the efficacy of ensemble learning in enhancing GNN capabilities for analyzing complex graph-structured data. 
    The code is public at \url{https://github. com/wongzhenhao/ELGNN}. 
\end{abstract}

\section{Introduction}

Graph is ubiquitous in a wide range of real-world relationships, ranging from social and rating networks~\cite{newman2002random} to biology networks~\cite{gilmer2017neural}. Currently, the Graph Neural Networks (GNNs)  have been proposed for learning representations over graph-structured
data. GNNs capture local graph structure and feature information via iterative aggregation of
features from neighbors using non-linear transformations. In this manner, GNNs have shown
promising performance on various applications on graph data. 

While GNNs can achieve fairly accurate results in numerous graph-based learning tasks, their enhanced ability to represent information comes at the cost of increased model complexity. This results in overfitting, which diminishes the model's capacity for generalization. The typical approach to addressing overfitting is Dropout~\cite{srivastava2014dropout}, a widely used regularization technique, but it may not be sufficient to prevent overfitting on its own. 

Given the \textit{ogbl-ddi} dataset, which is part of the Open Graph Benchmark~\cite{hu2020open} and represents a homogeneous, unweighted, undirected graph that models the network of drug-drug interactions, the primary goal is to prioritize true drug-drug interactions over false ones. In the context of the OGB leaderboard, existing methods predominantly revolve around Graph Neural Networks (GNNs) that capture characteristics of neighboring nodes. Some of these methods also incorporate edge features, like distance encoding, into their existing GNN frameworks. Additionally, a novel approach, PLNLP~\cite{wang2021pairwise}, builds upon GraphSage~\cite{hamilton2017inductive} and introduces a pairwise loss function, a common technique in siamese networks for ranking problems, which leads to improved performance. Furthermore, models based on attention mechanisms, such as AGDN~\cite{sun2020adaptive}, have also demonstrated outstanding performance. 

Ensemble learning is a widely recognized approach that enhances the effectiveness of machine learning tasks through the amalgamation and adjustment of predictions made by multiple models~\cite{yue2023relation, dietterich2000ensemble}. The primary reason for the effectiveness of the ensemble method is that the training dataset lacks sufficient information to identify the best single learner. Additionally, the hypotheses being explored and the scoring functions used to represent various characteristics may not encompass the actual target function. 

To address the limitation of existing GNNs which do not accurately represent the characteristics of the data, 
we proposed an innovative method called Ensemble Learning for Graph Neural Networks (ELGNN). In our work, we demonstrated how different models approach and prioritize structured information and node information. We explored various models to find an optimal and balanced weight ratio to observe their effects. Additionally, we employed Tree-Structured Parzen Estimator (TPE) for parameter search. 

With the integration of TPE for parameter search, our model consistently outperforms the latest GNN models and heuristic methods and has secured the top position on the \textit{ogbl-ddi} leaderboard for link prediction. This highlights the effectiveness of our approach in addressing the limitations of existing GNNs and optimizing model performance. 

\section{Proposed Methods}

\subsection{Preliminaries}
Our framework, ELGNN, aims to enhance the weight allocation among a set of individual machine learning models, resulting in enhanced performance for link prediction. We begin by introducing fundamental concepts of graph neural networks for link prediction and reviewing the leading model techniques on the \textit{ogbl-ddi} dataset. Following that, we present our framework, ELGNN. 

\subsubsection{Notations}
Denoted an undirected graph $\mathcal{G}=(\mathcal{V, E})$ with N nodes and E edges, where $\mathcal{V}=\{v_1, v_2, v_3, . . . , v_N\}$ and $\mathcal{E}= \{e_{ij}|v_i, v_j \in \mathcal{V}\}$. Within the graph, there are two types of information: structural information, which outlines how nodes are connected, and node feature information, which characterizes the attributes of the nodes. The connections between edges can be denoted using an adjacency matrix, $\mathnormal{A}\in \{0, 1\}^{N\times N}$. Let the initial node feature matrix $\mathcal{X}=\mathbb{R}^{N\times d^{(0)}}$, $\mathcal{X}=\{x_1, x_2, x_3, . . . x_N\}$, where every $x_i\in \mathcal{X}$ represents the d dimension node feature of $v_i$. 

\subsubsection{Graph Neural Networks for Link Prediction}
Formally, the \textit{l}-th layer in each GNN is defined as follows:
\begin{equation}
    \textit{h}^{(\textit{l})}=\sigma(A_{GNN}H^{(\textit{l}-1)}W^{(\textit{l}-1)}), 
\end{equation}
where the $A_{GNN}\in \mathbb{R}^{N\times N}$ is the normalized adjacency matrix depending on each GNN architecture. Generally, the widely used GNN architecture are GCN~\cite{kipf2016semi}, GAT~\cite{velivckovic2017graph}, SEAL~\cite{zhang2020revisiting}, Graphsage~\cite{hamilton2017inductive}. 
For example, $W^{(\textit{l})}\in \mathbb{R}^{d^{(\textit{l})}\times d^{(\textit{l}+1)}}$ is a trainable weight matrix meanwhile $\sigma (\cdot)$ is an activate function, such as Softmax, ReLU, and Sigmoid,
subsequently employed to forecast the presence of a link between node i and node j. 
After completing L layers of message passing, the node representations $H^{(L)}$ subsequently employed to forecast the presence of a link between node \textit{i} and node \textit{j}:

\begin{equation}
    \hat{\textit{y}}_{\textit{i}, \textit{j}}=\sigma(\textit{s}({\textit{h}_i}^{(L)}, {\textit{h}_j}^{(L)})), 
\end{equation}

where $\textit{s}(\cdot, \cdot)$ represents a link scoring function, usually MLP, and ${\textit{h}_i}^{(L)}$ is the representation of the node i from $H^{(L)}$. 

\subsubsection{Embedding}
The rapid adoption of embedding techniques following the introduction of word2vec has highlighted the critical importance of efficient embedding operations in influencing model performance. Embedding is the process of transforming complex graph data into low-dimensional vectors. It overcomes limitations in machine learning directly on the original graph and provides a more flexible and efficient computational approach. Furthermore, embedding effectively compresses large-scale graph data, avoiding the need to handle massive adjacency matrices. However, embedding must satisfy requirements such as attribute selection, scalability, and dimensionality selection to ensure an accurate description of graph topology and attributes, and to be suitable for large-scale networks. In this way, embedding enhances the performance and feasibility of applications with graph data. 

In general, embeddings are categorized into node embedding and graph embedding. Classic methods for node embedding include DeepWalk~\cite{perozzi2014deepwalk}, Node2vec~\cite{grover2016node2vec}, and SDNE~\cite{wang2016structural}. Graph embedding condenses the entire graph into a single vector, as seen in methods like Graph2vec~\cite{narayanan2017graph2vec}, which employs the skip-gram concept to convert the entire graph into a vector space. 

\subsection{Representative Methods on OGB}
While it may appear that employing a greater number of learner members would result in enhanced performance, Zhou et al. established the "many could be better than all" theorem, challenging this notion~\cite{zhou2002ensembling}. Opting for the choice of assembling an ensemble with a few effective models rather than utilizing all of them is generally a more favorable decision. In this subsection, we provide a concise overview of some top strategies on \textit{ogbl-ddi} dataset.

\subsubsection{Adaptive Graph Diffusion Network}
Chuxiong Sun et al.~\cite{sun2020adaptive} proposed an Adaptive Graph Diffusion Network (AGDN) which executes multi-layer generalized graph diffusion across various feature spaces with reasonable computational complexity and runtime. Typical graph diffusion techniques utilize extensive powers of the transition matrix along with predefined weighting coefficients. In contrast, AGDNs merge smaller multi-hop node representations with adaptable and generalized weighting coefficients. They propose two scalable mechanisms of weighting coefficients, Hop-wise Attention (HA) and Hop-wise Convolution (HC), to capture multi-hop information in a graph. They believe that AGDN can outperform other methods based on structural features due to the high density of the \textit{ogbl-ddi} dataset, rendering structural patterns almost irrelevant. 

\subsubsection{Graph Inception Diffusion Network}
Graph Inception Diffusion Network (GIDN)~\cite{wang2022gidn} extends graph diffusion across diverse feature spaces and employs the inception module to mitigate the substantial computational load arising from intricate network architectures. "Inception" refers to a neural network architecture module that was popularized by Google's Inception models~\cite{szegedy2015going}. It is designed to capture a wide range of features at various levels of abstraction within the network, which can be particularly useful for image recognition tasks. The module uses a combination of different filter sizes and pooling operations to process input data in a more efficient manner, making it suitable for deep neural networks. Notably, GIDN achieved top-1 performance even before the introduction of our model. 

\subsubsection{Path-aware Siamese GNN}
Different from AGDN, Path-aware Siamese Graph Neural Network (PSG) captures information related to both nodes and edge features for a pair of nodes. Specifically, it focuses on preserving the structure information of k-neighborhoods and gathering relay path information for the nodes~\cite{lv2022path}. Moreover, an innovative multi-task GNN framework is used, which incorporates self-supervised contrastive learning to distinguish between positive and negative links, all while concurrently capturing both the content and behavior of nodes. 

\subsection{ELGNN}
In order to amalgamate the member learners into more effective and robust learners, we propose the ELGNN. We selected several top OGB models not only for their outstanding performance but also because they are designed based on different approaches. We believe that choosing either structural information or node information alone would result in information loss. Therefore, we construct an ensemble of graph neural networks by utilizing both input features and graph structures to find the optimal balance ratio for \textit{ogbl-ddi}. 

\begin{figure*}[htbp]
    \centering
    \includegraphics[width=1\linewidth]{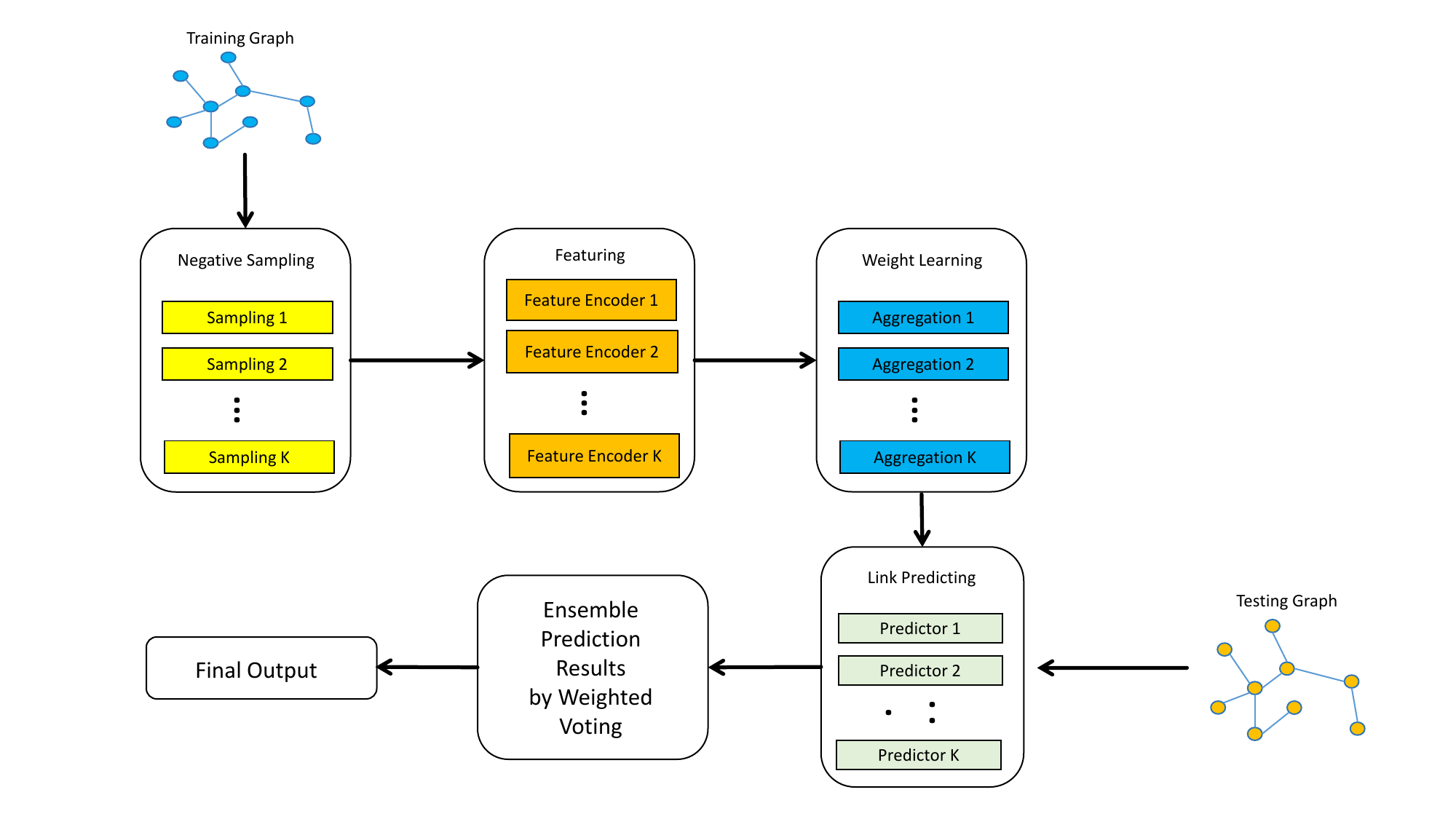}
    \caption{An overview of ELGNN architecture. 
    For the transductive setting, the input training graph and input test graph form the full graph. 
    The K represents the number of model ensembles, for here, K=3. }
    \label{fig:model}
\end{figure*}

 Our goal is to have the model rank true drug interactions higher than non-interacting drug pairs. More specifically, we rank each true drug interaction within a set of sampled negative drug interactions. In this regard, we solely focus on finding the optimal weights based on the positive samples. Furthermore, to preserve the distinctive characteristics of each model, we perform ensemble learning at the prediction layer of each model, thus learning enhanced prediction parameters, as shown in Figure~\ref{fig:model}. \
 
Our method adopts weighted voting, where the ultimate decision is determined by a weighted combination of individual model decisions. We assign weights to each model based on their prediction scores. For a given positive node pair (\textit{i}, \textit{j}), the objective function is as follows:
\begin{equation}
\begin{aligned}
    \hat{\textit{y}}_{\textit{i}, \textit{j}}
    = \max\nolimits_{ \{ \alpha^k \} }(\sum\nolimits_{ k } \alpha^k{\hat{\textit{y}}_{\textit{i}, \textit{j}}}^k), \\
     s. t. \quad \sum{\alpha^k}=1, 
\end{aligned}
\end{equation}
where the \textit{k} represents the model we selected for ensemble learning, which is AGDN, GIDN, and PSG. By harnessing the scores indicating link presence within these prediction regions, the objective function acquires the optimal score for link presence associated with a given input, considering both node information and structural information. In the context of hyperparameter optimization, our method employs the Tree-structured Parzen Estimator (TPE) for an efficient, adaptive, and automated exploration of hyperparameter space. Tree-Structured Parzen Estimator (TPE) is a Bayesian optimization technique commonly used for hyperparameter tuning and optimizing machine learning models~\cite{bergstra2011algorithms}. TPE builds a probabilistic model of the objective function and uses it to efficiently explore the hyperparameter space. It balances exploration and exploitation, making it particularly useful for finding the best hyperparameters for a given model. Therefore, ELGNN is an effective method to improve the model performance. 

\section{Experiments}
In this section, we will delve into a detailed discussion and analysis of the performance of the proposed ELGNN method. We have carried out extensive experiments to validate the superiority of our approach by assessing the improvement in prediction accuracy. 
\subsection{Experimental Setup}

\subsubsection{Dataset}
    We evaluate our ELGNN on the Open Graph Benchmark dataset, \textit{ogbl-ddi}. 
    The \textit{ogbl-ddi} dataset is a homogeneous, unweighted, undirected graph, 
    representing the drug-drug interaction network~\cite{wishart2018drugbank}. 
    Each node in the network represents an FDA-approved or experimental drug. 
    The edges symbolize interactions between these drugs, 
    indicating situations where the combined effect of taking two drugs together significantly deviates from the expected effect when the drugs act independently of each other. 
    The prediction task involves forecasting drug-drug interactions based on available data regarding previously known drug-drug interactions. 
    Statistics of the dataset are provided in Table~\ref{tab:dataset}.

\begin{table}[ht]
    \centering
    \caption{Dataset statistics}
    \small
    \begin{tabular}{c|ccc}
    \toprule
         Data Split & Nodes & Positive Edges & Negative Egdes \\ \midrule
         Train     & 4267 & 1067911 & -- \\ 
         Validation & --    & 133489   & 101882 \\
         Testing & -- & 133489  &95599 \\
    \bottomrule
    \end{tabular}
    \label{tab:dataset}
\end{table}

\subsubsection{Evalution Metric}
OGB provides standardized dataset splits and evaluators that allow for easy and reliable comparison of different models in a unified manner. For \textit{ogbl-ddi} dataset, the evaluation metric is a ranking metric. More specifically, we rank each true drug interaction within a set of around 100, 000 randomly sampled negative drug interactions and calculate the ratio of positive interactions that are ranked at position K or higher. This metric is referred to as Hits@K, and for the \textit{ogbl-ddi} dataset, K is defined as 20.

\subsubsection{Settings}
The experiments are conducted under the circumstance of NVIDIA A100 GPU (80G RAM). 
In order to demonstrate the effectiveness of our approach, 
we used these models' codes which have been officially made public by OGB directly with their corresponding hyperparameters. 
It is summarized in Table~\ref{tab:settings}. 

\begin{table}[htbp]
    \centering
    \caption{Settings of models}
    \small
    \setlength\tabcolsep{2pt}
    \begin{tabular}{c|ccc}
    \toprule
        Methods & AGDN & GIDN & PSG \\
        \midrule
         Epoches & 1000 & 1000 & 500 \\
         Parameters & 3, 506, 691 & 3, 506, 691 & 3, 499, 009 \\
         Num. of GNN layers & 2 & 2 & 2 \\
         Num. of MLP layers & 2 & 2 & 2 \\
         Num. of neg. sampling & 3 & 3 & 3\\
         Learning rate & 0.001 & 0.003 & 0.001 \\
         Dim. of node emb. & 512 & 512 &512 \\
         Encoder & GAT & GAT & GraphSAGE \\
         Loss function & AUC loss & AUC loss & AUC loss \\
         \bottomrule
    \end{tabular}
    \label{tab:settings}
\end{table}

\subsection{Accuracy Improvement}
In this section, we demonstrate how our method incorporates various feature extraction and aggregation methods to enhance link prediction results. As shown in Table~\ref{tab:accuracy improved}, ELGNN beats the state-of-the-art algorithm GIDN on the leaderboard until submission and achieves 2. 3\% more performance improvement in terms of the Hits@20 Test. It indicates our ensemble learning method can significantly enhance the graph representation learning of GNN networks, and hence can enhance the prediction effectiveness of link prediction-based GNN models for dataset \textit{ogbl-ddi}. 

\begin{table}
    \centering
    \caption{Link Prediction Performance of our method and other GNN models on dataset \textit{ogbl-ddi}. 
    \textbf{Bold} indicates the second best performance. }
    \small
    \setlength\tabcolsep{2pt}
    \begin{tabular}{c|cc}
    \toprule
        Methods & Valid. Hits@20 & Test Hits@20 \\
        \midrule
         GCN & 0.5550 ± 0.0208& 0.3707 ± 0.0507 \\
         GraphSAGE &0.6262 ± 0.0037 & 0.5390 ± 0.0474 \\
         GCN+JKNet & 0.6776 ± 0.0095 & 0.6056 ± 0.0869 \\
         GraphSAGE + Edge Attr & 0.8044 ± 0.0404 & 0.8781 ± 0.0474 \\
         PLNLP & 0.8242 ± 0.0253	& 0.9088 ± 0.0313 \\
         PSG & 0.8306 ± 0.0134 & 0.9284 ± 0.0047 \\
         AGDN & 0.8943 ± 0.0281 & 0.9538 ± 0.0094 \\
         GIDN & 0.8258 ± 0.0000 & 0.9542 ± 0.0000 \\ \midrule
         ELGNN(Ours) & \textbf{0.8965 ± 0.0021} & \textbf{0.9777 ± 0.0037} \\
         \bottomrule
    \end{tabular}
    \label{tab:accuracy improved}
\end{table}

\subsection{Ablation Study}
We additionally perform an ablation study to illustrate the efficacy of ELGNN when utilizing TPE for link prediction tasks. We present the outcomes for both the baseline approach (with weight averaging) and ELGNN (optimized with TPE) under identical settings on the \textit{ogbl-ddi} dataset. 
As indicated in Table ~\ref{tab:ablation study}, TPE notably attains a 2.2\% increase in performance compared to the baseline, thereby enhancing the link prediction performance to state-of-the-art levels. This underscores that our method is effective for link prediction tasks. 
\begin{table}[H]
    \centering
    \caption{Averaging vs. TPE}
    \begin{tabular}{c|cc}
    \toprule
         Method&  Valid. & Test \\
         \midrule
         Averaging& 0.8549 &0.9574 \\
         TPE & 0.9095 & 0.9796 \\
        \bottomrule
    \end{tabular}
    \label{tab:ablation study}
\end{table}

\section{Conclusion}
In this paper, we introduce ELGNN, an ensemble learning technique designed to create a combination of random decision GNNs. ELGNN focuses on consolidating and refining predictions from multiple models, with its primary goal being to optimize the prediction score for link presence in a given input, accommodating a variety of feature extraction and aggregation approaches. What makes ELGNN unique is its efficient parallelization, allowing each base model to be independently trained and make predictions. Furthermore, this method enables the use of a diverse set of GNN models as the foundation for the ensemble. Extensive experimental results on real-world benchmark datasets consistently show that ELGNN surpasses all baseline methods, setting a new standard for link prediction performance on the \textit{ogbl-ddi} dataset. As part of future research, we intend to further enhance ELGNN by incorporating additional heuristic methods for link prediction and extending its applicability to various datasets. 

\paragraph{Limitations.}
The importance of the base model cannot be directly explained by the ensemble weights due to the difference in prediction score scales. However, this will not affect the final result when TPE is used to search the ensemble weights.

 \section*{Author Contributions Statement}
Z. Wong contributes to idea development, algorithm implementation, experimental design, result analysis, and paper writing. L. Yue contributes to idea development, experimental design, result analysis, and paper writing. Q. Yao contributes to idea development. All authors read, edited, and approved the paper.

\clearpage
\bibliography{anthology, custom}
\bibliographystyle{acl_natbib}

\end{document}